# Modeling Object Recognition in Newborn Chicks using Deep Neural Networks

**Donsuk Lee (donslee@iu.edu)[1,2], Denizhan Pak (denpak@iu.edu)[1], Justin N. Wood (woodjn@iu.edu)[1-4]**

[1]Department of Informatics,
[2]Cognitive Science Program,
[3]Center for the Integrated Study of Animal Behavior,
[4]Department of Neuroscience,
Indiana University, Bloomington, IN 47408 USA

### Abstract

In recent years, the brain and cognitive sciences have made great strides developing a mechanistic understanding of object recognition in mature brains. Despite this progress, fundamental questions remain about the origins and computational foundations of object recognition. What learning algorithms underlie object recognition in newborn brains? Since newborn animals learn largely through unsupervised learning, we explored whether unsupervised learning algorithms can be used to predict the view-invariant object recognition behavior of newborn chicks. Specifically, we used feature representations derived from unsupervised deep neural networks (DNNs) as inputs to cognitive models of categorization. We show that features derived from unsupervised DNNs make competitive predictions about chick behavior compared to supervised features. More generally, we argue that linking controlled-rearing studies to image-computable DNN models opens new experimental avenues for studying the origins and computational basis of object recognition in newborn animals.

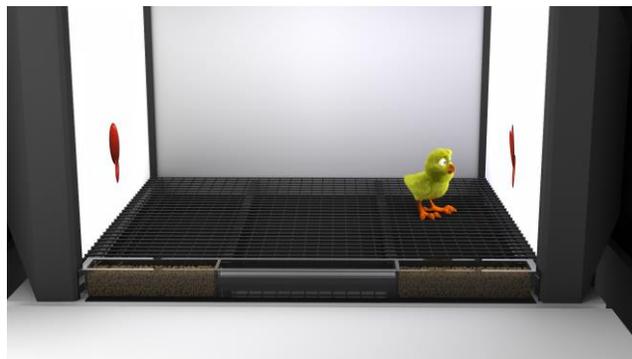

Figure 1: Illustration of the controlled-rearing chambers. The chambers contained no objects other than the virtual objects projected on the display walls. During the input phase, the chicks were exposed to a single virtual object (imprinted object). During the test phase, the imprinted object was projected on one display wall and an unfamiliar object was projected on the opposite display wall, in a two-alternative forced choice test.

## Introduction

One of the great unsolved mysteries in cognitive science concerns the origins and computational basis of object recognition. What mechanisms underlie object recognition in newborn brains, and how are those mechanisms shaped by experience? To date, the vast majority of studies have used representations learned by deep neural network (DNN) models to predict behavioral and neural responses in *mature* visual systems (Bashivan, Kar, & DiCarlo, 2019; Battleday, Peterson, & Griffiths, 2020; Cadieu et al., 2014; Kriegeskorte, 2015; Schrimpf et al., 2020). Such DNN models are typically trained with supervised learning, in which the model learns to categorize objects from thousands to millions of labeled training images. The feature representations of DNN models trained with supervised learning can accurately predict behavioral and neural responses to novel images of objects in humans and nonhuman primates (Battleday et al., 2020; Yamins & DiCarlo, 2016). Furthermore, DNN models can be used to control the activity state of entire populations of neural sites (Bashivan et al., 2019). DNN models thus provide a mechanistic understanding of object recognition in mature visual systems.

However, supervised DNN models do not accurately describe how object recognition emerges and develops in biological brains. Unlike supervised DNN models, both mature animals and newborn animals learn largely through unsupervised learning. Human infants start receiving labeled object input only when they begin understanding language, and nonhuman animals receive little (if any) labeled training input during development. Understanding the origins of object recognition therefore requires building computational models that learn like newborn brains, using unsupervised learning algorithms.

Recently, researchers have made significant progress building unsupervised learning systems that can perform well on challenging object recognition tasks (e.g. Image Net; Deng et al., 2009). Moreover, one particular class of unsupervised DNN models—deep unsupervised contrastive embedding methods—can achieve neural prediction accuracy in multiple areas of the ventral visual system that equals or exceeds supervised methods (Zhuang et al., 2021). These methods produce brain-like representations even when trained solely on head-mounted camera data collected from young children, suggesting that these unsupervised DNNs can serve as biologically plausible hypotheses of learning in developing brains. To date, however, it has not been possible to apply computational modeling approaches to understand how *newborn* brains learn about the world.

There are two major barriers in building computational models of object recognition in newborn brains. First, newborn subjects are hard to study. Most methods for studying development in newborn subjects are low powered and produce noisy measurements of behavior (Wood & Wood, 2019). This makes it challenging to obtain the precise benchmarks needed to build accurate computational models. Second, it is not possible to control the environment in which most newborn animals are raised. Thus, for most animals, we do not know which visual experiences were used to 'train' their visual system during development. Since the outputs of DNN models change radically as a function of the training data, an accurate comparison of DNN models and newborn animals requires training the models and animals with the same set of visual experiences.

To overcome these barriers, Wood (2013) developed an automated controlled-rearing method (described in the next section). This method allows researchers to strictly control the visual experiences of newborn animals (newly hatched chicks) and obtain precise behavioral benchmarks of each chick's object recognition performance (Figure 1). Since the chicks are reared in controlled environments, we can simulate (reproduce) all of the visual training data available to the chicks, and then give that same training data to computational models. By training animals and models on the same set of training data, we can directly compare their learning abilities.

Here, we investigated whether feature representations derived from unsupervised DNNs can be used to model object recognition (categorization) behavior of newborn chicks in one of these automated controlled-rearing studies (Wood, 2013). If the learning mechanisms in DNN models resemble those used by newborn brains, then it should be possible to use the features learned by DNN models to predict chicks' recognition behavior. Importantly, the DNN models were trained using the same visual experiences presented to the newborn chicks.

Specifically, we trained DNNs with unsupervised objective functions and used the images of the objects that the newborn chicks received during the controlled-rearing experiment. Then, we used object representations extracted from the networks as inputs for cognitive models of categorization (modified versions of prototype and exemplar models) in order to predict the behavior of the chicks. We show that the categorization models operating over unsupervised DNN feature representations make competitive predictions compared to the features derived from supervised DNNs. These results suggest that combining controlled-rearing experiments, DNNs, and cognitive models can be a promising research direction for studying the origins and computational basis of object recognition in newborn brains.

## Automated Controlled-Rearing Studies of Newborn Chicks

Reverse engineering the origins of object recognition requires precise benchmarks showing how specific visual inputs shape the development of object recognition. To obtain these benchmarks, we use newborn chicks as a model system. Unlike most animals, chicks are uniquely suited for studying the earliest stages of visual development. Chicks are mobile on the first day of life, require no parental care, and can be raised in strictly controlled environments immediately after hatching. With chicks, it is therefore possible to study how experience shapes the earliest stages of postnatal visual development.

Results from previous studies suggest that many object perception abilities emerge rapidly during development; for instance, newborn chicks are capable of visual parsing (Wood & Wood, 2021), visual binding (Wood, 2014), view-invariant object recognition (Wood, 2013, 2015; Wood & Wood, 2020; Wood & Wood, 2015), rapid object recognition (Wood & Wood, 2017), and object permanence (Prasad, Wood, & Wood, 2019). Remarkably, all of these abilities emerge when chicks are reared with a single object, indicating that newborn brains are capable of "one-shot" visual learning. From an artificial intelligence perspective, this is an impressive feat. Machine learning systems typically require thousands to millions of labeled training images to develop object recognition, whereas chicks develop object recognition from input of a single object seen from a limited range of views.

## Modeling Object Recognition Behavior of Newborn Chicks

In this work, we focus on modeling the behavioral results from Wood (2013). In the study, newborn chicks were raised for 2 weeks in controlled-rearing chambers. In the first week, the chicks' visual experience was limited to a single virtual object rotating through a 60° viewpoint range. In the second week, the chicks were tested on their ability to recognize that object from novel viewpoints, using a two-alternative forced choice test. The chicks succeeded on the task, generating view-invariant representations that generalized across large, novel, and complex changes in the object's appearance.

### Task Definition

We first provide a formal definition of the two-alternative forced choice task that was used to measure view-invariant object recognition in newborn chicks (Wood, 2013). In the task, a visually-naive subject receives a set of $k$ examples $X = \{x_1, ..., x_k\}$ from a *single* object category during the Input (Imprinting) phase. Specifically, X consists of images of the imprinted object rotating through a 60° viewpoint range. Note that, unlike in traditional categorization tasks, the examples of the second category are not provided during the Input phase. During each test trial, the chick is then shown a pair of stimuli $\mathbf{y} = (y^+, y^-)$, where $y^+$ and $y^-$ are images of objects from the familiar (positive) and novel (negative) categories, respectively. The set of test stimuli consisted of familiar and novel images presented across a variety of viewpoint ranges. The task is to decide which of the two test stimuli belongs to the same category as the learned examples in X. Task performance was evaluated based on the percentage of time the chicks spent with the imprinted object versus the unfamiliar object. If the chicks developed view-invariant object recognition, then they should have spent more time with the imprinted object than the unfamiliar object.

## Categorization Models for Two-Alternative Forced Choice Task

Our categorization models for the two-alternative forced choice task are derived from the prototype (Rosch, 1973) and exemplar (Nosofsky, 1986) theories of categorization. Both theories have been formalized as quantitative models and widely used to study human (McKinley & Nosofsky, 1995) and animal (Smith, Zakrzewski, Johnson, Valleau, & Church, 2016) categorization behaviors. While there is a long-standing debate about whether categories are represented by exemplars or by prototypes, it is not our primary goal to provide evidence for one or the other in this study. Rather, our goal was to explore whether either model could predict performance in this task when combined with features derived from unsupervised DNNs.

In the case of typical 2-way categorization tasks, the categorization models express the probability that an input stimulus is classified into Category A as follows:

$$p(\text{Choose A} \mid y) = \frac{\text{Sim}(y, c_A)^\gamma}{\text{Sim}(y, c_A)^\gamma + \text{Sim}(y, c_B)^\gamma},$$

where $\text{Sim}(\cdot)$ is some function that measures the similarity of an input stimulus $y$ to the category representation $c_A$ or $c_B$ of either category A or B. $\gamma$ is a free parameter which determines the steepness of the decision boundary. Prototype and exemplar models differ in the types of similarity functions they use. In prototype models, category abstractions (usually the average of the members in each category) are used for comparison, whereas in exemplar models, all of the learned examples are stored and used for comparison.

In our two-alternative forced choice variant of prototype and exemplar models, the probability of choosing the positive object $y^+$ (familiar category) given the learned single-category examples X and a pair of test stimuli $\mathbf{y} = (y^+, y^-)$ is expressed as:

$$p(\text{Choose } y^+ \mid y^+, y^-) = \frac{\text{Sim}(y^+, c_X)^\gamma}{\text{Sim}(y^+, c_X)^\gamma + \text{Sim}(y^-, c_X)^\gamma}.$$

In the above expression, $c_X$ is the representation of the familiar category constructed by examples in X. The choice of similarity functions for prototype and exemplar models are described below.

In the prototype model, the familiar (imprinted) category is represented by the average of members in X:

$$c_X = \frac{1}{|X|} \sum_{x \in X} x.$$

Then, the similarity function is an exponential function of the negative distance D between a stimulus y and the category representation $c_X$:

$$\text{Sim}(y, c_X) = e^{-D(y, c_X)},$$

In the exemplar model, the familiar category is represented by all of the members in X, i.e. $c_X = X$. Then, the similarity function for the exemplar model is the summation of the distances between a stimulus y and each exemplar $x \in c_X$:

$$\text{Sim}(y, c_X) = \sum_{x \in c_X} e^{-\beta D(y, x)},$$

where $\beta$ is a free model parameter that determines the rate at which similarity declines with distance.

In both exemplar and prototype models, we used squared Mahalanobis distance with diagonal covariance matrix $I\sigma$ to express their similarity functions:

$$D(y, x) = \sum_{i=1}^{d} \sigma_i |x_i - y_i|^2.$$

Here, x, y and σ are vectors in *d*-dimensional space. In exemplar models, $\sigma_i$ is often referred to as attention weights (McKinley & Nosofsky, 1995). When $\sigma_i = 1$, the set of points equidistant from a given location is a sphere. With $\sigma_i$ as free parameters, the models have flexibility to stretch this sphere to correct the respective scales of feature dimensions.

## Learning Feature Representations Using Unsupervised Deep Neural Networks

Typically, psychological models of categorization use hand-coded descriptions or features derived from similarity judgements to represent stimuli. In the real world, however, brains must convert high-dimensional sensory inputs ($10^6$ optic nerve fibers) into object-centric representations that can guide adaptive behavior. Given that early visual experience can heavily shape object recognition (e.g., Wood & Wood, 2016; 2018), we argue that the features that serve as inputs to these cognitive models should be learned from visual experience. We take a first step towards this goal by using feature representations learned by DNNs.

We used a similar approach as Battleday, Peterson, & Griffiths (2020), extracting features from DNNs to serve as inputs to the categorization models. We then compared the features extracted from supervised and unsupervised DNNs in terms of their ability to predict the chicks' performance. Crucially, we trained our DNN feature extractors using images sampled from *virtual* controlled-rearing chambers that were designed to mimic the controlled-rearing chambers in which we collected the data from the newborn chicks.

The key principle of unsupervised DNN models is that self-supervised "proxy" tasks (e.g. reconstructing the input image) can produce representations that are useful for downstream tasks (e.g. object recognition). We used two classes of unsupervised learning algorithms to train our DNN feature extractors.

**Variational autoencoder.** The variational autoencoder (Kingma & Welling, 2014) is a class of unsupervised generative models that approximates the true distribution of the observed data using variational objective functions. It has been proposed that a modification (β-VAE; Higgins et al., 2017) to the original objective function of VAEs can result in "disentangled" visual representations, which is critical for biological visual systems to solve invariant object recognition tasks (DiCarlo & Cox, 2007). $\beta$-VAE models are trained to maximize the following objective:

$$\mathbb{E}_{p(\mathbf{x})}\left[\mathbb{E}_{q_\phi(\mathbf{z}|\mathbf{x})}[\log p_\theta(\mathbf{x}|\mathbf{z})] - \beta \mathrm{KL}\left(q_\phi(\mathbf{z}|\mathbf{x}) \parallel p(\mathbf{z})\right)\right],$$

where $\mathbf{z}$ is the latent factor generating data $\mathbf{x}$, and $p_\theta(\mathbf{x}|\mathbf{z})$ and $q_\phi(\mathbf{z}|\mathbf{x})$ are learned by the encoder and decoder neural networks. This objective with $\beta = 1$ corresponds to the original formulation by Kingma & Welling (2014). The hyperparameter $\beta$ is set to a value greater than 1 in order to learn disentangled representations $\mathbf{z}$.

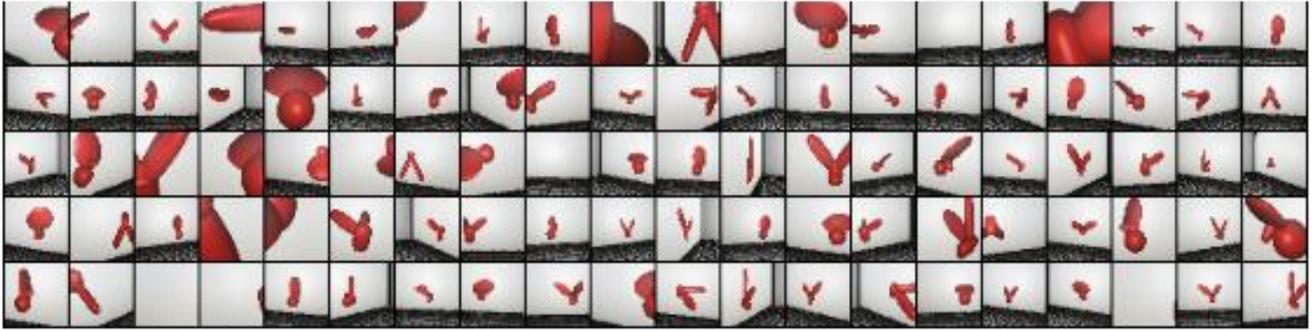

Figure 2: Sample images from the synthetic dataset used to train the DNN feature extractors. The dataset was constructed by recording the visual observations of an agent moving within a virtual controlled-rearing chamber.

**Contrastive learning.** In addition, we used a contrastive learning method entitled SimCLR (Chen, Kornblith, Norouzi, & Hinton, 2020). Contrastive algorithms rely on self-supervised prediction tasks for learning discriminative representations. These methods learn an embedding space by maximizing the distance between the "dissimilar" (negative) samples while minimizing the distance between "similar" (positive) samples. Unlike supervised methods, contrastive learning methods require an unsupervised approach to characterize positive and negative examples. The SimCLR (Chen et al., 2020) creates a positive pair by applying random image transformations such as color distortion, rotation, or Gaussian blur to each image $x$. A DNN encoder is then used to produce representations for the transformed image pairs. The contrastive loss is calculated to minimize the distance between the representations of positive pairs, while maximizing their distance to other samples.

## Method

**Stimuli.** The stimulus set from Wood (2013) consisted of 24 animations of two 3D objects (12 animations for each object). Each animation displayed an object rotating through a 60° viewpoint range about an axis passing through its centroid, completing the full back and forth rotation every 6s. For each object animation, we sampled 26 frames spanning the 60° rotation (2.3° rotation in 0.12s between two adjacent frames), so the final stimulus set consisted of 624 images. Each of the sampled images was treated as a single input stimulus for the categorization models.

**Chick behavioral data.** We used the behavioral data collected by Wood (2013), which included 35 newborn chicks. After hatching, each chick was moved from the incubator to a controlled-rearing chamber in darkness. The controlled-rearing chamber was equipped with two monitors that were used to display the object animations. The chicks' entire visual object experience was limited to the virtual objects projected on the monitors. In the Input phase, the imprinted object was displayed from a single 60° viewpoint range. Half of the chicks were imprinted to object A, and half of the chicks were imprinted to object B. In the Test phase, the chicks were tested with 12 viewpoint ranges of familiar and unfamiliar objects. Test trials were scored as "correct" when the chicks spent a greater proportion of time with their imprinted object and "incorrect" when the chicks spent a greater proportion of time with the unfamiliar object.

**DNN architectures and training.** In order to mimic the visual experiences of the chicks raised in the controlled-rearing chambers, we constructed an image dataset by sampling visual observations of an agent that moved randomly within a virtual controlled-rearing chamber (Figure 2). The virtual chamber and agent were created with the Unity 3D game engine and the ML-Agents Toolkit (Juliani et al., 2020). We programmatically removed images that did not contain an object. The resulting dataset contained 240,000 images (10k images for each of the 24 object animations). All DNN feature extractors were trained on this synthetic dataset.

We used a standard ResNet architecture (He, Zhang, Ren, & Sun, 2016) with 18 layers as the base encoder for all of our DNN feature extractors. For $\beta$-VAE, the encoder was followed by two fully-connected heads with 256 units, which inferred Gaussian mean and variance of the posterior latent distribution. The decoder architecture was in the reverse order of the encoder architecture. We trained three $\beta$-VAEs with $\beta = 0, 1,$ and 4. For the SimCLR model, the encoder was followed by a 2-layer MLP projection head. All DNNs were trained and optimized using the Adam optimization algorithm (Kingma & Ba, 2017) for 100 epochs.

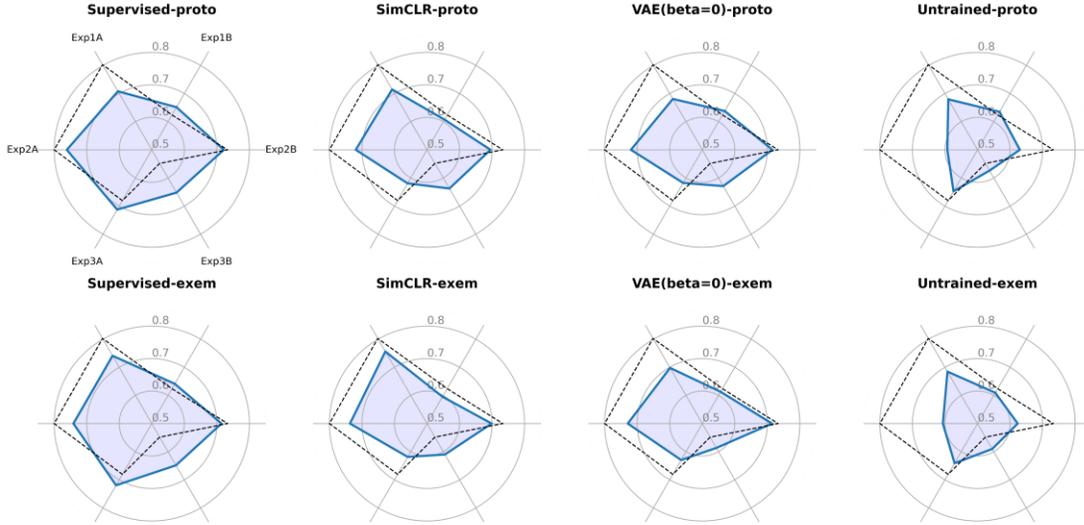

Figure 3: Actual task performance of the chicks (dotted lines) and predicted task performance of the models (blue lines). Each axis represents the average task accuracy across each of the six experimental conditions (Exp. 1A/B, 2A/B, & 3A/B, where A and B denote each of the two imprinted objects). The graphs show the predictions for each of the four feature representations (supervised, SimCLR, VAE, & untrained) in the prototype (proto) and exemplar (exem) categorization models.

**Feature representations.** We extracted features for each stimulus using DNNs trained with the unsupervised objectives described in the previous section. For both SimCLR and $\beta$-VAE, the features were taken from the last layer of the ResNet base encoder. The feature dimensions were 512.

**Categorization model fitting.** We fitted the categorization models by maximizing the log-likelihood of the observed behavioral data. We used the Adam algorithm (Kingma & Ba, 2017) with a learning rate of 0.003 and a batch size of 256 to find the parameters of the categorization models. We performed a 6-fold cross validation on the chick behavioral data by randomly splitting the data by test conditions. Model parameters were chosen based on the average log-likelihood on held-out data across the 6 validation folds. We used a PyTorch package to implement and optimize all categorization models.

**Model comparison.** We used the cross-validated negative log-likelihood (NLL) averaged across folds to evaluate goodness of fit (lower NLL indicates better fit). In addition, we calculated the Pearson correlation between the model predictions and the task accuracies of the chicks across the test conditions to measure how much behavioral variance is captured by each model.

We estimated the noise ceiling by computing the split-half correlation of the behavioral data. We randomly split the subjects' responses for each pair of test stimuli into two half-sets, computed the correlation between the two sets, and applied a Spearman-Brown correction. We repeated these steps 100 times and averaged the corrected correlations to obtain the final estimate.

**Baselines.** We had two baseline features: **untrained** and **supervised**. The untrained features were extracted from a randomly initialized ResNet-18 encoder. For the supervised features, we trained a ResNet-18 encoder followed by a fully-connected classification layer. The supervised ResNet was trained with the same image dataset as the unsupervised feature extractors, but critically, the model was optimized for binary cross-entropy loss using ground truth category labels. We expected that supervised features would make accurate predictions about chicks' categorization behavior because they compactly encode the category structure of the stimuli.

## Results

Table 1 shows the NLL and correlation scores for 12 different combinations of DNN features and categorization models. First, the results show that prototype and exemplar models produce comparable results for all feature representations. While exemplar models outperformed prototype models with most of the feature representations, the differences were not large. As discussed in Battleday et al. (2020), this pattern suggests that the structure of the categories might be relatively simple in relation to the dimensionality of features we used in our experiments.

Next, we compared the best model scores across the feature representations (bold numbers in Table 1). As expected, untrained features resulted in poor fits and supervised features produced the best fit out of all feature representations. Also, all unsupervised features outperformed untrained features. Notably, the contrastive features (SimCLR) performed similar to the supervised features, with only 0.003 difference in NLL compared to supervised features. Contrary to our expectation, $\beta$-VAE features performed the best with the lowest disentanglement parameter ($\beta = 0$), in which case the $\beta$-VAE reduces to a basic convolutional autoencoder.

While all unsupervised features obtained slightly lower NLL scores compared to the supervised features, SimCLR

Table 1: Model results. Lower NLL scores, and higher correlation scores, indicate better fit to the chick data.

| Features | Model | NLL | Corr. |
|---|---|---|---|
| Untrained | *Prototype* | **0.640** | **0.407** |
|  | *Exemplar* | 0.641 | 0.442 |
| Supervised | *Prototype* | 0.608 | 0.464 |
|  | *Exemplar* | **0.606** | **0.543** |
| VAE ($\beta$=0) | *Prototype* | 0.622 | 0.398 |
|  | *Exemplar* | **0.616** | **0.561** |
| VAE ($\beta$=1) | *Prototype* | 0.675 | -0.190 |
|  | *Exemplar* | **0.673** | **-0.152** |
| VAE ($\beta$=4) | *Prototype* | **0.693** | **0.0** |
|  | *Exemplar* | 0.693 | 0.0 |
| SimCLR | *Prototype* | 0.625 | 0.449 |
|  | *Exemplar* | **0.609** | **0.649** |
| * *Correlation noise ceiling* |  | - | 0.753 |

and VAE ($\beta$=0) features achieved higher correlation scores than the supervised features. This implies that these unsupervised features can outperform the supervised features in accounting for variance in newborn chicks' categorization behavior. In Figure 3, we display the model predictions and the chicks' categorization performance across the six experimental conditions from Wood (2013). The supervised features often overestimated the chicks' performance (in Exp1B, 3A, 3B) because the supervised encoder learned highly discriminative features. Conversely, unsupervised features mostly underestimated the chicks' performance but could better capture their confusion patterns.

In general, these results indicate that unsupervised methods can build representational spaces that predict chicks' object recognition performance, without the need for the biologically unrealistic labelling of training data required for supervised methods.

## Discussion

Psychological models of categorization have produced formal and predictive models of categorization behavior in human adults. Here we build on this rich research tradition by linking cognitive models of object categorization to controlled-rearing studies of newborn animals. Specifically, we explored whether cognitive models can accurately predict the emerging object recognition behavior of newborn chicks, within natural settings that involve complex, high-dimensional stimuli. We found that when cognitive models are provided with input features derived from supervised and unsupervised DNNs, the cognitive models could predict chicks' behavioral performance.

One benefit of using DNNs is that they are image computable, meaning that they can generate responses for arbitrary input images. By using DNNs to learn high-dimensional feature spaces, we can make comparisons across animals and models without making prior assumptions about the specific features used to build object representations in newborn brains. Additionally, the fact that unsupervised features perform comparably to supervised features indicates that we can use self-supervised learning algorithms to develop computational theories of visual learning, using behavioral data from newborn animals as benchmarks.

There are, of course, many more unsupervised learning models that could be explored within this framework. There is also much more work that needs to be done to build a better understanding of the feature spaces produced by DNN models. The approach described here—which involves linking controlled-rearing studies of newborn animals to computational models in artificial intelligence—may prove helpful in this regard. By training newborn animals and DNNs with the same set of high-dimensional sensory data, we can control a potentially large source of variation in the learning of visual features. Ultimately, we argue that this approach provides an experimental avenue for reverse engineering the learning mechanisms in newborn brains and discovering the algorithmic principles that drive intelligent behavior.

## Limitations and Future Work

There are at least two limitations to the present work. First, although controlled-rearing studies allow us to impose strong constraints on the training data provided to DNNs, there are still some mismatches in the training data presented to the chicks and the DNNs that prohibit direct comparison between newborn animals and computational models. Besides some obvious mismatches like input resolution and field of view, our current approach ignores the temporal order of the visual input. Moreover, to make this modeling approach more tractable, we assumed that the visual inputs from all of the test stimuli were available to the chicks during learning. In the experiment, however, the chicks were only presented with a single viewpoint range during the input phase. In future work, we will reduce the amount of training data available to the DNNs to more precisely match the training data provided to the newborn chicks during the imprinting period.

Second, the objective functions of contrastive learning algorithms may not be (entirely) biologically plausible. As far as we know, there is no biological evidence that the visual system employs stochastic transformations of sensory inputs (e.g. random color distortion) for learning visual representations. A promising future direction is to use more biologically plausible unsupervised learning algorithms that exploit spatiotemporal information. For instance, recent studies suggest that deep spatiotemporal contrastive learning methods can achieve primate-level representation learning (Zhuang et al., 2021). In future work, we will use a similar approach to explore whether contrastive learning methods can build the same kinds of object representations as newborn chicks when provided with similar streams of high-dimensional sensory data.

## Acknowledgements

Funded by NSF CAREER Grant BCS-1351892 and a James S. McDonnell Foundation Understanding Human Cognition Scholar Award.